\documentclass[sigconf]{acmart}

\usepackage{booktabs} 

\usepackage{graphicx}
\usepackage{subcaption}
\usepackage{amsmath}
\usepackage[linesnumbered,ruled]{algorithm2e}

\usepackage{capt-of}
\usepackage{booktabs}
\usepackage{varwidth}
\newsavebox\tmpbox

\newcommand{\nop}[1]{}
\newcommand{\acret}{\textbf{ACRET}}

\setcopyright{none}

\renewcommand\footnotetextcopyrightpermission[1]{} 
\begin{document}
\title{Accelerating Dependency Graph Learning from Heterogeneous Categorical Event Streams via Knowledge Transfer}


\author{Chen Luo}
\authornote{The work was done when the first author was on an internship at NECLA.}
\affiliation{%
  \institution{Rice University}
  \city{Houston} 
  \state{Texas} 
}
\email{cl67@rice.edu}

\author{Zhengzhang Chen}
\affiliation{%
  \institution{NEC Labs America}
  \city{Princeton} 
  \state{New Jersey} 
}
\email{zchen@nec-labs.com}

\author{Lu-An Tang}
\affiliation{%
  \institution{NEC Labs America}
  \city{Princeton} 
  \state{New Jersey} 
}
\email{ltang@nec-labs.com}

\author{Anshumali Shrivastava}
\affiliation{%
  \institution{Rice University}
  \city{Houston} 
  \state{Texas} 
}
\email{anshumali@rice.edu}

\author{Zhichun Li}
\affiliation{%
  \institution{NEC Labs America}
  \city{Princeton} 
  \state{New Jersey} 
}
\email{zhichun@nec-labs.com}


\begin{abstract}
Dependency graph, as a heterogeneous graph representing the intrinsic relationships between different pairs of system entities, is essential to many data analysis applications, such as root cause diagnosis, intrusion detection, \textit{etc}. Given a well-trained dependency graph from a source domain and an immature dependency graph from a target domain, how can we extract the entity and dependency knowledge from the source to enhance the target? One way is to directly apply a mature dependency graph learned from a source domain to the target domain. But due to the domain variety problem, directly using the source dependency graph often can not achieve good performance. Traditional transfer learning methods mainly focus on numerical data and are not applicable.

In this paper, we propose \acret, a knowledge transfer based model for accelerating dependency graph learning from heterogeneous categorical event streams. In particular, we first propose an entity estimation model to\nop{estimate the probability of each source domain entity that can be included in the final dependency graph of the target domain. by utilizing source domain heterogeneous relations from the categorical event streams. Then, we propose a domain adaptation model to construct the dependency relationships for the target domain graph.} filter out irrelevant entities from the source domain based on entity embedding and manifold learning. Only the entities with statistically high correlations are transferred to the target domain. On the surviving entities, we propose a dependency construction model for constructing the unbiased dependency relationships by solving a two-constraint optimization problem.  The experimental results on synthetic and real-world datasets demonstrate the effectiveness and efficiency of \acret. 
We also apply \acret\ to a real enterprise security system for intrusion detection. 
Our method is able to achieve superior detection performance at least 20 days lead lag time in advance with more than 70\% accuracy.
\end{abstract}





\maketitle

\section{Introduction}

The heterogeneous categorical event data are ubiquitous. 
Consider system surveillance data in enterprise networks, where each data point is a system event that involves heterogeneous types of entities: time, user, source process, destination process, and so on. 
Mining such event data is a challenging task due to the unique characteristics of the data: (1) the exponentially large event space. For example, in a typical enterprise network, hundreds (or thousands) of hosts incessantly generate operational data. A single host normally generates more than $10,000$ events per second; And (2) the data varieties and dynamics. The variety of system entity types may necessitate high-dimensional features in subsequent processing, and the event data may changing dramatically over time, especially considering the heterogeneous categorical event streams \cite{aggarwal2003framework,aggarwal2007data,manzoor2016fast}. 

\begin{figure*}[!htbp]
\centering
\includegraphics[width=\textwidth]{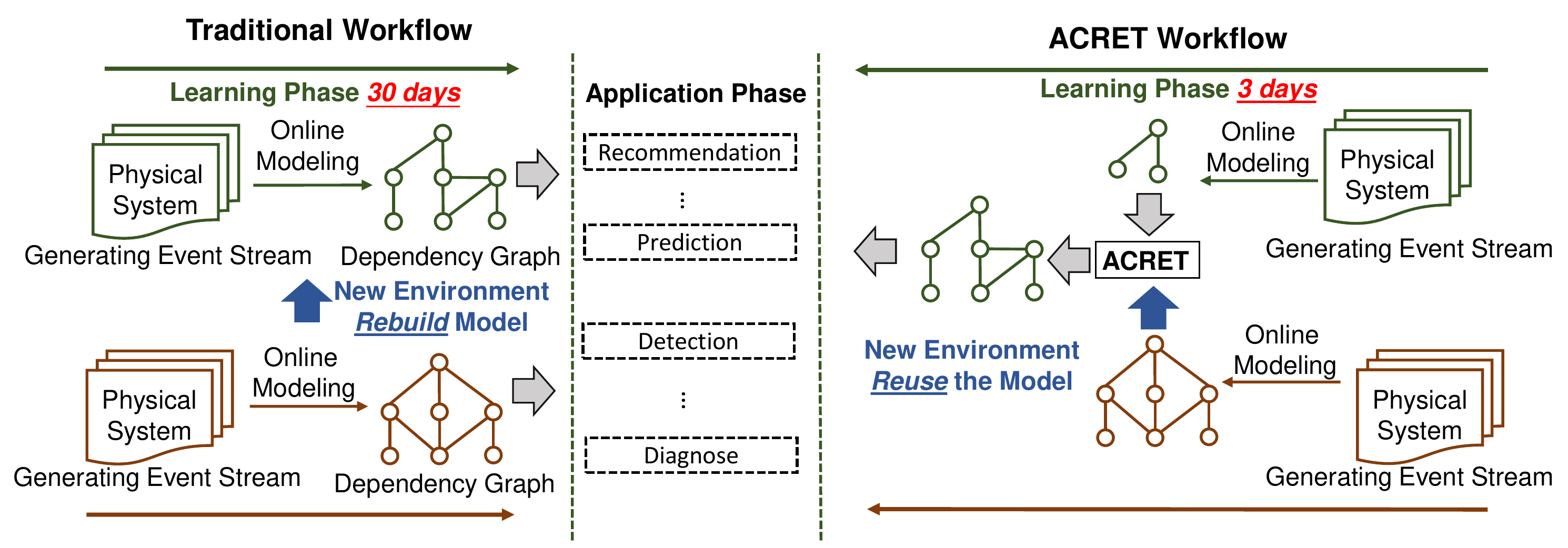}
\caption{Comparison between a traditional work flow and \acret\ work flow of learning the dependency graph. \acret\ extracts the knowledge from a well-trained dependency graph to speed up the training process of a new dependency graph.}
\label{fig:workflow} 
\end{figure*}

To address the above challenges, the recent studies of dependency graphs \cite{he2011dependency,king2005enriching,xu2016high} have witnessed a growing interest. Such dependency graphs can be applied to model a variety of systems including enterprise networks \cite{xu2016high}, societies~\cite{myers2014information}, ecosystems~\cite{kawale2013graph}, \textit{etc}. For instance, we can present an enterprise network as a dependency graph, with nodes representing system entities of processes, files, or network sockets, and edges representing the system events between entities (\textit{e.g.}, a process reads a file). 
This enterprise system dependency graph can be applied to many forensic analysis tasks such as intrusion detection, risk analysis, and root cause diagnosis \cite{xu2016high}.
A social network can also be modeled as a dependency graph representing the social interactions between different users.
Then, this social dependency graph can be used for user behavior analysis or abnormal user detection~\cite{kim2011modeling}.

 

However, due to the aforementioned data characteristics, learning a mature dependency graph from heterogeneous categorical event streams often requires a long period of time. 
For instance, the dependency graph of an enterprise network needs to be trained for several weeks before it can be applied for intrusion detection or risk analysis as illustrated in Fig. \ref{fig:workflow}.
Furthermore, every time, when the system is deployed in a new environment, we need to rebuild the entire dependency graph. 
This process is both time and resource consuming.

Enlightened by the cloud services \cite{krutz2010cloud}, 
one way to avoid the time-consuming rebuilding process is by reusing a unified dependency graph model in different domains/environments.
However, due to the domain variety, directly apply the dependency graph learned from an old domain to a new domain often can not achieve good performance. 
For example, the enterprise network from an IT company (active environment) is very different from the enterprise network from an electric company (stable environment). Thus, the enterprise dependency graph of the IT company contains many unique system entities that can not be found in the dependency graph of the electric company. 
Nevertheless, there are still a lot of room for transfer learning.
\nop{Directly deploy the model learned from one environment to a new environment will suffered from the domain differences. }
\nop{Our experiment in Section \ref{sec:exp} demonstrate the impracticable of direct reuse.}

Transfer learning has shed light on how to tackle the domain differences \cite{pan2010survey}.
It has been successfully applied in various data mining and machine learning tasks, such as clustering and classification \cite{chen2015net2net}.
However, most of the transfer learning algorithms focus on numerical data \cite{dai2007boosting,sun2015transfer,chattopadhyay2013joint}. 
When it comes to graph structure data, there is less existing work~\cite{fang2015trgraph,he2009graph}, not to mention the dependency graph. 
This motivates us to propose a novel knowledge transfer-based method for dependency graph learning.

In this paper, we propose \acret, a knowledge transfer based method for \underline{a}\underline{c}cele\underline{r}ating d\underline{e}pendency graph learning from heterogeneous ca\underline{t}egorical event streams.
\acret\ consists of two sub-models: \textbf{EEM} (Entity Estimation Model) and \textbf{DCM} (Dependency Construction Model).
Specifically, first, \textbf{EEM} filters out irrelevant entities from source domain based on entity embedding and manifold learning.
Only the entities with statistically high correlations can be transferred to the target domain.
Then, based on the reduced entities, \textbf{DCM} model effectively constructs unbiased dependency relationships between different entities for the target dependency graph by solving a two-constraint optimization problem. We launch an extensive set of experiments on both synthetic and real-world data to evaluate the performance of \acret.
The results demonstrate the effectiveness and efficiency of our proposed algorithm. 
We also apply \acret\ to a real enterprise security system for intrusion detection.
Our method is able to achieve superior detection performance at least 20 days lead lag time in advance with more than 70\% accuracy.

\section{Preliminaries and Problem Statement}
\label{sec:pre}

\begin{figure*}[t]
    \centering
     \includegraphics[width=\textwidth]{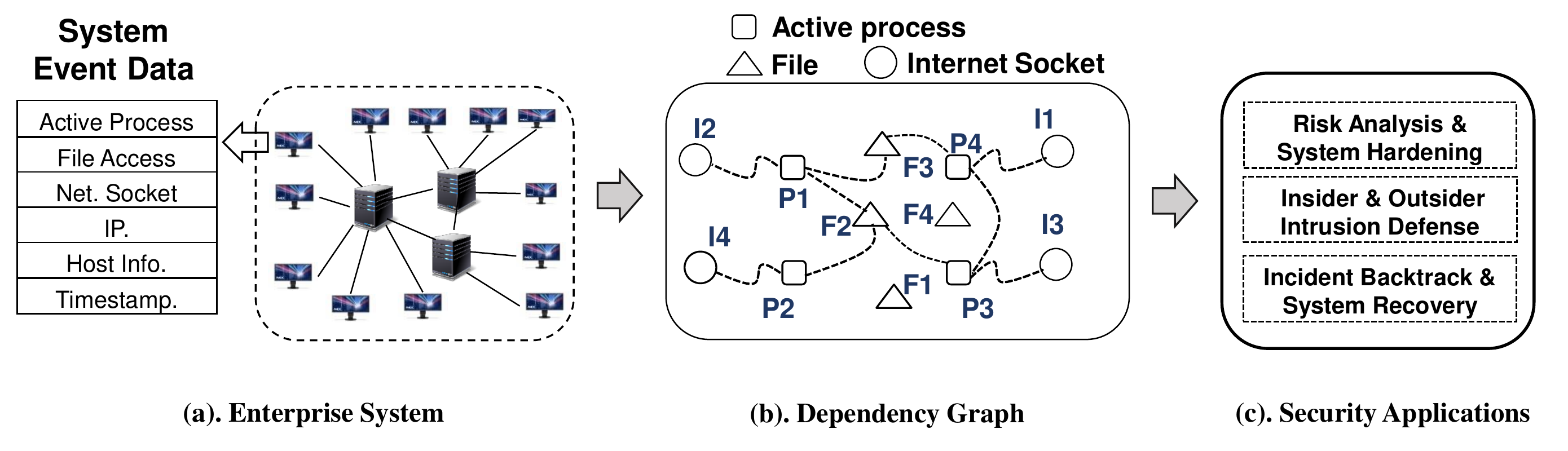}
      \caption{An overview of an enterprise security system.}
    \label{fig:motivating}
\end{figure*}

In this section, we introduce some notations and define the problem.

\textbf{Heterogeneous Categorical Event}.
A heterogeneous categorical event $e = (a_1, · · · , a_m)$ is a record contains $m$ different categorical attributes, and the $i$-th attribute value $a_i$ denotes an entity from the type $\mathcal{T}_i$. 

For example, in the enterprise system (as illustrated in Fig. \ref{fig:motivating}), a process event (\textit{e.g.}, a program opens a file or connects to a server) can be regarded as a heterogeneous categorical event.
It contains information, such as timing, type of operation, information flow directions, user, and source/destination process, \textit{etc}.

By continuous monitoring/auditing the heterogeneous categorical event data (streams) generated by the physical system, one can generate the corresponding dependency graph of the system, as in~\cite{he2011dependency,king2005enriching,xu2016high}.
This dependency graph is a heterogeneous graph representing the dependencies/interactions between different pairs of entities. Formally, we define the dependency graph as follows:




\noindent{\bf Dependency Graph.}
A dependency graph is a heterogeneous undirected weighted graph $G = \{V, E\}$, 
where $V = \{v_0, v_1, ..., v_n\}$ is the set of heterogeneous system entities, and $n$ is the total number of entities in the dependency graph; $E = \{e_0, e_1, ..., e_m\}$ is the set of dependency relationships/edges between different entities. For ease of discussion, we use the terms edge and dependency interchangeably in this paper. A undirected edge $e_i(v_k, v_j)$ between a pair of entities $v_k$ and $v_j$ exists depending on whether they have a dependency relation or not. The weight of the edge denotes the intensity of the dependency relation. \nop{A undirected edge $e_i(v_s, v_d)$ between a pair of entities $v_s$ and $v_d$ exists depending on whether the they have a causality relation or not.
The weight for the dependency denotes the intensity of the causality relation.}

In an enterprise system, a dependency graph can be a weighted graph between different system entities, such as processes, files, users, Internet sockets.
The edges in the dependency graph are the causality relations between different entities. 

As shown in Fig. \ref{fig:motivating}, the enterprise security system utilities the accumulated historical heterogeneous system data from event streams to construct the system dependency graph and update the graph periodically. The learned dependency graph is applied to forensic analysis applications such as intrusion detection, risk analysis, and incident backtrack \textit{etc}.

\nop{However, learning a dependency graph often requires a system auditing for a long period of time which in turn causes overwhelmingly large amount of system audit events. 
A system dependency graph needs to be trained for more than three weeks before deployed to the real world anomaly detection systems. 
In addition, when we deploy the security system to a new environment. 
We often need to spend other several weeks to retrain the dependency graph.}



The problems of cold-start and time-consuming training reflect a great demand for an automated tool for effectively transferring dependency graphs between different domains.
Motivated by this, this paper focuses on accelerating the dependency graph learning via knowledge transfer. Based on the definitions described above, we formally define our problem as follows:

\textbf{Knowledge Transfer for Dependency Graph Learning}.
Given two domains: a source domain $\mathcal{D}_S$ and a target domain $\mathcal{D}_T$. 
In the source domain $\mathcal{D}_S$, we have a well-trained dependency graph $G_S$ generated from the heterogeneous categorical event streams. 
In target domain $\mathcal{D}_T$, we have a small incomplete dependency graph $\widehat{G_T}$ trained by a short period of time.
The task of knowledge transfer for dependency graph learning is to use $G_S$ to help construct a mature dependency graph $G_T$ in the domain $\mathcal{D}_T$. 
 
There are two major assumptions for this problem: (1) The event streams in the source domain and target domain are generated by the same physical system; (2) The entity size of source dependency graph $G_S$ should be larger than the size of the intersection graph $G_S \cap \widehat{G_T}$. Because transferring knowledge from a less informative dependency graph to an informative graph is unreasonable.
 
\nop{For example, $\mathcal{D}_S$ can be an enterprise system environment from an IT company, and $\mathcal{D}_T$ can be another enterprise system environment from electronic management servers.
In next section, we present a motivation example for this problem.}

\nop{However, there is no existing approach that can be directly used to solve this problem. This motivates us to design a new approach to effectively and efficiently transferring dependency graphs from different domains. In the next section, we will introduce the details of our proposed transferring framework.}


\begin{figure}[!htbp]
    \centering
    \includegraphics[width=0.45\textwidth]{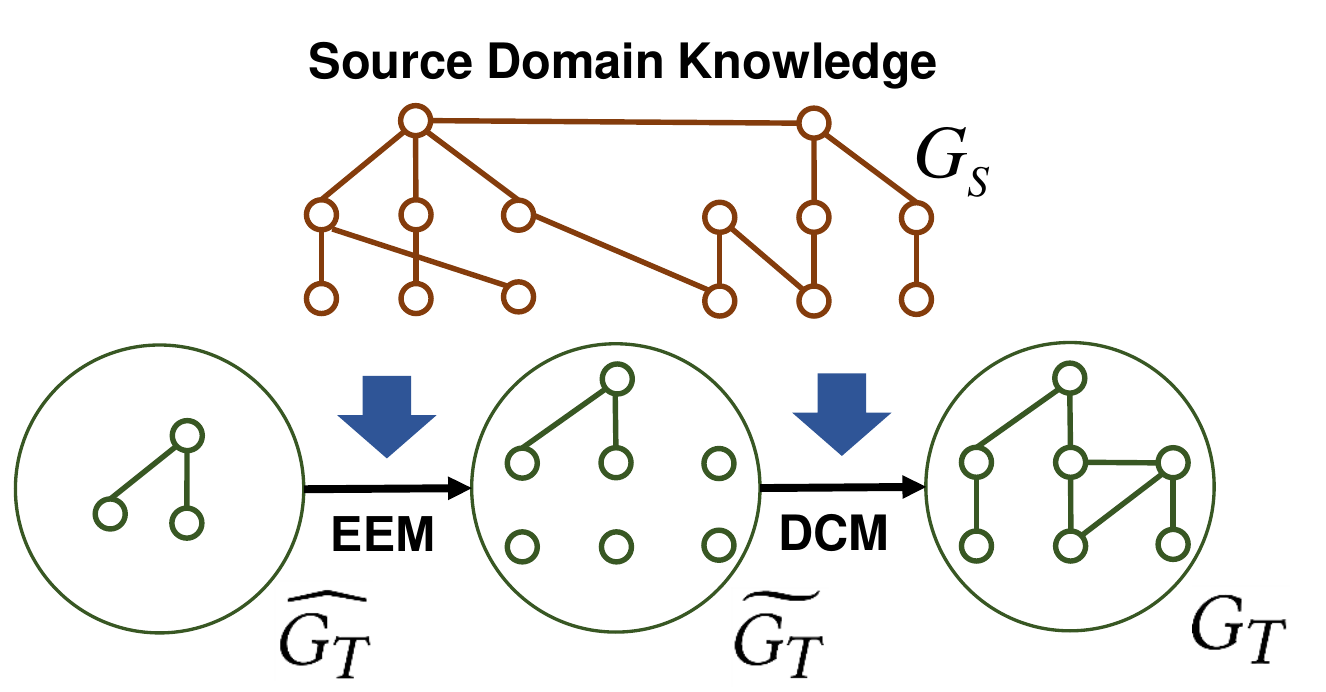}
    \caption{An overview of the \acret\ model.}
    \label{fig:frame} 
\end{figure}

\section{The \acret\ Model}

\nop{In this section, we introduce the proposed model in details. }
\nop{We now describe the \acret\ model in details.
As we introduced in section \ref{sec:pre}, a dependency graph consists of two basic elements: (1) entities and (2) dependencies between entities.
So we design two sub-models for estimating each element respectively.
These two models are: EEM (Entity Estimation Model) and DCM (Dependency Construction Model) as illustrated in Fig. \ref{fig:frame}.
We first introduce these two sub-models in details separately and then combine them into a uniform algorithm.}
To learn a mature dependency graph $G_T$, intuitively, we would like to leverage the entity and dependency information from the well-trained source dependency graph $G_S$ to help complete the original small dependency graph $\widehat{G_T}$. One naive way is to directly transfer all the entities and dependencies from the source domain to the target domain. However, due to the domain difference, it is likely that there are many entities and their corresponding dependencies that appear in source domain but not in the target domain. Thus, one key challenge in our problem is how to identify the domain-specific/irrelevant entities from the source dependency graph. After removing the irrelevant entities, another challenge is how to construct the dependencies between the transferred entities by adapting the domain difference and following the same dependency structure as in $\widehat{G_T}$. To address these two key challenges in dependency graph learning, we propose a knowledge transfer algorithm with two sub-models: \textbf{EEM} (Entity Estimation Model) and \textbf{DCM} (Dependency Construction Model) as illustrated in Fig.~\ref{fig:frame}. We first introduce these two sub-models separately in details, and then combine them into a uniform algorithm.
\subsection{\textbf{EEM}: \nop{Embedding-based} Entity Estimation Model}
\label{sec:entity}

For the first sub-model, Entity Estimation Model, our goal is to filter out the entities in the source dependency graph ${G}_S$ that are irrelevant to the target domain. 
To achieve this, we need to deal with two main challenges: (1) the lack of intrinsic correlation measures among categorical entities, and (2) heterogeneous relations among different entities in the dependency graph.

To overcome the lack of intrinsic correlation measures among categorical entities, we embed entities into a common latent space where their semantics can be preserved. 
More specifically, each entity, such as a user, or a process in computer systems, is represented as a $d$-dimensional vector and will be automatically learned from the data. 
In the embedding space, the correlation of entities can be naturally computed by distance/similarity measures in the space, such as Euclidean distances, vector dot product, and so on. Compared with other distance/similarity metrics defined on sets, such as Jaccard similarity, the embedding method is more flexible and it has nice properties such as transitivity \cite{zhang2015categorical}.

To address the challenge of heterogeneous relations among different entities, we use the \textit{meta-path} proposed in \cite{sun2012mining} to model the heterogeneous relations.
For example, in a computer system, a \textit{meta-path} can be a 
``Process-File-Process", or a "File-Process-Internet Socket".
``Process-File-Process" denotes the relationship of two processes load the same file, and "File-Process-Internet Socket" denotes the relationship of a file loaded by a process who opened an Internet Socket.

The potential \textit{meta-path}s induced from the heterogeneous network $G_S$ can be infinite, but not every one is relevant and useful for the specific task of interest. There are some works \cite{chen2017task} for automatically selecting the \textit{meta-path}s for specific tasks.
\nop{For more details about \textit{meta-path} selecting methods, please refer \cite{chen2017task}.}





Given a set of \textit{meta-path}s $P = \{ p_1, p_2, ...\}$, where $p_i$ denotes the $i$-th \textit{meta-path} and let $|P|$ denotes the number of \textit{meta-path}s.
We can construct $|P|$ graphs $G_{p_i}$ by each time only extracting the corresponding \textit{meta-path} $p_i$ from the dependency graph~\cite{sun2012mining}. Let $u_S$ denotes the vector representation of the entities in $G_S$.
Then, we model the relationship between two entities $u_S(i)$ and $u_S(j)$ as:
\begin{equation}
\label{equ:ref}
\left \|u_S(i) - u_S(j)\right \|_F^2 \approx S_{G}(i,j),
\end{equation}
In the above, $S_{G}$ is a weighted average of all the similarity matrices $S_{p_i}$: 
\begin{equation}
\label{equ:graphs}
S_{G} = \sum_{i=1}^{|P|} w_i S_{p_i}, 
\end{equation}
where $w_i$'s are non-negative coefficients, and $S_{p_i}$ is the similarity matrix constructed by calculating the pairwise shortest path between each entities in $A_{p_i}$.
Here, $A_{p_i}$ is the adjacent matrix of the dependency graph $G_{p_i}$.
By using the shortest path in the graph, one can capture the long term relationship between different entities~\cite{bondy1976graph}.
Putting Eq. \ref{equ:graphs} into Eq. \ref{equ:ref}, we have:
\begin{equation}
\begin{split}
\left \|u_S(i) - u_S(j)\right \|_F^2 & \approx\sum_{i=1}^{|P|} w_i S_{p_i},
\end{split}
\end{equation}
where $\left \|*\right \|_F^2$ is is the Frobenius norm \cite{han2011data}.

Then, the objective function of \textbf{EEM} model is:
\begin{equation}
 \mathcal{L}_1 ^{(u_S, W)} = \sum_{i,j}^{n} \left( \left \|u_S(i) - u_S(j)\right \|_F^2 - S_{G} \right)^{\theta} + \Omega (u_S, W), 
\end{equation}
where $W = \{w_1, w_2, ..., w_{|P|}\}$, and $\Omega (u_S, W) = \lambda \left \| u_S \right\| + \lambda \left \| W \right\|$ is the generalization term \cite{han2011data}, which prevents the model from over-fitting.
$\lambda$ is the trade-off factor of the generalization term. 
In practice, we can choose $\theta$ as 1 or 2, which bears the resemblance to Hamming distance and Euclidean distance, respectively.

Putting everything together, we get:
\begin{equation}
\label{equ:l1}
\begin{split}
 \mathcal{L}_1 ^{(u_S, W)} & = \sum_{i,j}^{n} \left(  \left \|u_S(i) - u_S(j)\right \|_F^2 - S_{G} \right)^{\theta}  + \Omega (u_S, W)\\
&= \sum_{i,j}^{n} \left(  \left \|u_S(i) - u_S(j)\right \|_F^2 - \sum_{i=0}^{|P|-1} w_i S_{p_i} \right)^{\theta}  + \lambda \left \| u_S \right\| + \lambda \left \| W \right\|
\end{split}
\end{equation}

Then, the optimized value $\{u_S, W \}^{opt}$ can be obtained by:
\[
\{u_S, W\}^{opt} = \arg \min_{u_S, W} \mathcal{L}_1 ^{(u_S, W)}.
\]

\subsubsection{Inference Method}
The objective function in Eq. \ref{equ:l1} contains two sets of parameters: (1) $u_S$, and (2) $W$.
Then, we propose a two-step iterative method for optimizing $ \mathcal{L}_1 ^{(u_S, W)}$, where the entity vector matrices $u_S$ and the weight for each \textit{meta-path} $W$ mutually enhance each other. 
In the first step, we fix the weight vectors $W$ and learn the best entity vector matrix $u_S$. In the second step, we fix the entity vector matrix $u_S$ and learn the best weight vectors $W$.

\textbf{Fix $W$ and learn $u_S$:}
when we fix $W$, then the problem is reduced to $\left \|u_S(i) - u_S(j)\right \|_F^2 \approx S_{G}(i,j)$, where $S_{G}$ is a constant similarity matrix.
Then, the optimization process becomes a traditional manifold learning problem.
Fortunately, we can have a closed form to solve this problem, via so called multi-dimensional scaling \cite{han2011data}.
To obtain such an embedding, we compute the eigenvalue decomposition of the following matrix:
\[- \frac{1}{2}HS_{G}H = U\Lambda U,\] where $H$ is the double centering matrix, $U$ has columns as the eigenvectors and $\Lambda$ is a diagonal matrix with eigenvalues. Then, the embedding $u_S$ can be chosen as:
\begin{equation}
\label{equ:es}
u_S = U_k\sqrt[2]{\Lambda_k}.
\end{equation}

\textbf{Fix $u_S$ and learn $W$}:
When fixing $u_S$, the problem is reduced to:
\begin{equation}
\begin{split}
 \mathcal{L}_1 ^{W} &= \sum_{i,j}^{n} \left( \left \|u_S(i) - u_S(j)\right \|_F^2 - \sum_{i=1}^{|P|} w_i S_{p_i} \right)^{\theta} + \lambda \left \| u_S \right\| + \lambda \left \| W \right\| \\
 &= \sum_{i,j}^{n} \left( C_1 - \sum_{i=0}^{|P|} w_i S_{p_i} \right)^{\theta} + \lambda \left \| W \right\| + C_2,
\end{split}
\end{equation}
where $C_1 = \left \|u_S(i) - u_S(j)\right \|_F^2$ is a constant matrix, and $C_2 = \lambda \left \| E_S \right\|$ is also a constant.
Then, this function becomes a linear regression. 
So, we also have the close form solution for $W$:
\[
W = (S_G^TS_G)^{-1}S_GC_1.
\]

After we get the embedding vectors $u_S$, then the relevance matrix $\mathbb{R}$ between different entities can be obtained as:
\begin{equation}
\mathbb{R} = u_Su_S^T
\end{equation}

One can use a user defined threshold to select the entities with high correlation with target domain for transferring.
But user defined threshold is often suffered by the lack of domain knowledge. 
So here, we introduce a hypothesis test based method for automatically thresholding the selection of the entities.

For each entity in $\widehat{G_T}$, we first normalize all the scores by:
$\mathbb{R}(i,:)_{norm} = (\mathbb{R}(i,:) - \mu)/{\delta}$, where $\mu = \overline{\mathbb{R}(i,:)}$ is the average value of $\mathbb{R}(i,:)$, and $\delta$ is the standard deviation of $\mathbb{R}(i,:)$. This standardized scores can be approximated with a gaussian distribution.
Then, the threshold will be $1.96$ with $P=0.025$. (or $2.58$ for $P=0.001$) \cite{han2011data}.
By using this threshold, one can filter out all the statistically irrelevant entities from the source domain, and transfer highly correlated entities to the target domain.


By combining the transferred entities and the original target domain dependency graph $\widehat{G_T}$, we get $\widetilde{G_T}$, as shown in Fig. \ref{fig:frame}.
Then, the next step is to construct the missing dependencies in $\widetilde{G_T}$.

\subsection{DCM: Dependency Construction Model}
\label{sec:link}


To construct the missing dependencies/edges in $\widetilde{G_T}$, there are two constraints need to be considered:
\begin{itemize}
\item \textit{Smoothness Constraint}: The predicted dependency structure in $G_T$ needs to be close to the dependency structure of the original graph $\widetilde{G_T}$.
The intuition behind this constraint is that the learned dependencies should more or less intact in $\widetilde{G_T}$ as much as possible.
\item \textit{Consistency Constraint}: 
Inconsistency between $\widetilde{G_T}$ and $\widetilde{G}_S$ should be similar to the inconsistency between $\widehat{G_T}$ and $\widehat{G_S}$.
Here, $\widetilde{G_S}$ and $\widehat{G_S}$ are the sub-graphs of $G_S$ which have the same entity set with $\widetilde{G_T}$ and $\widehat{G_T}$, respectively.
This constraint guarantees that the target graph learned by our model can keep the original domain difference with the source graph. 
\end{itemize}

Before we model the above two constraints, we first need a measure to evaluate the inconsistence between different domains.
In this work, we propose a novel metric named dynamic factor $F(\widetilde{G_S},\widetilde{G_T})$ between two dependency graphs $\widetilde{G_S}$ and $\widetilde{G_T}$ from two different domains as:
\begin{equation}
\label{equ:df}
\begin{split}
F(\widetilde{G_S},\widetilde{G_T}) &= \frac{\left\| \widetilde{A_S} - \widetilde{A_T}\right\|}{ |\widetilde{G_S}|*(|\widetilde{G_S}| - 1) / 2} = \frac{2\left\| \widetilde{A_S} - \widetilde{A_T}\right\|}{ n_S(n_S-1)},
\end{split}
\end{equation}
where $n_s = |\widetilde{G_S}|$ is the number of entities in $\widetilde{G_S}$, $\widetilde{A_S}$ and $\widetilde{A_T}$ denote the adjacent matrix of $\widetilde{G_S}$ and $\widetilde{G_T}$, respectively, and $n_S(n_S-1) / 2$ denotes the number of edges of a fully connected graph with $n_S$ entities \cite{bondy1976graph}. 

Next, we introduce the Dependency Construction Model in details.

\subsubsection{Modeling Smoothness Constraint}
\label{sec:mcc}
We first model the smoothness constraint as follows:
\begin{equation}
\label{equ:l21}
\begin{split}
 \mathcal{L}_{2.1} ^{u_T} & = \left \| \sum_{i=1}^{n_S} \sum_{j=0}^{n_S-1} \left(u_T(i)u_t(j)^T - \widetilde{A_T}(i,j) \right) \right \|_F^2 + \lambda \left \| u_T \right\| \\
&= \left \|u_Tu_T^T - \widetilde{A_T}\right \|_F^2 + \Omega(u_T),
\end{split}
\end{equation}
where $u_T$ is the vector representation of the entities in $G_T$, and $\Omega(u_T) = \lambda \left \| u_T \right\|$ is the regularization term. 

\subsubsection{Modeling Consistency Constraint}
\label{sec:msc}
We then model the consistency constraint as follows:
\begin{equation}
\label{equ:l22}
\begin{split}
 \mathcal{L}_{2.2} ^{(u_T)} & = 
 \left \|F(u_Tu_T^T, \widetilde{A_S}) - F(\widehat{A_S},\widehat{A_T}) \right \|_F^2 + \Omega(u_T),
\end{split}
\end{equation}
where $F(*,*)$ is the dynamic factor as we defined before. Then, putting Eq. \ref{equ:df} and $\Omega(u_T)$ into Eq. \ref{equ:l22}, we get:

\begin{equation}
\begin{split}
\mathcal{L}_{2.2} ^{E_T} & = \left \|F(u_Tu_T^T, \widetilde{G_S}) - F(\widehat{G_S},\widehat{G_T}) \right \|_F^2 + \Omega(u_T) \\
& = \left \|\frac{2\left\| u_Tu_T^T - \widetilde{A_S}\right\|}{ n_s(n_S-1)} - F(\widehat{G_S},\widehat{G_T}) \right \|_F^2 + \Omega(u_T) \\
& = \left \|\frac{2\left\| u_Tu_T^T - \widetilde{A_S}\right\|}{ n_s(n_s-1)} - C_3 \right \|_F^2 + \Omega(u_T),
\end{split}
\end{equation}
where $C_3 = F(\widehat{G_S},\widehat{G_T})$.

\subsubsection{Unified Model}
Having proposed the modeling approaches in Section \ref{sec:mcc} and \ref{sec:msc}, we intend to put all the two constraints together.
The unified model for dependency construction is proposed as follows:
\begin{equation}
\label{equ:l2}
\begin{split}
\mathcal{L}_{2} ^{u_T} & = \mu \mathcal{L}_{2.1} ^{u_T} + (1 - \mu)\mathcal{L}_{2.2} ^{u_T} \\
& = \mu \left \|u_Tu_T^T - \widetilde{A_T}\right \|_F^2 
 + (1 - \mu)\left \|\frac{2\left\| u_Tu_T^T - \widetilde{A_S}\right\|}{ n_S(n_S-1)} - C_3 \right \|_F^2 + \Omega(u_T) 
\end{split}
\end{equation}

\nop{In the above, the symbols have the same meanings as introduced in previous sections. }
The first term of the model incorporates the \textit{Smoothness Constraint} component, which keeps the $u_T$ closer to target domain knowledge existed in the $\widetilde{G}_S$. 
The second term considers the \textit{Consistency Constraint}, that is the inconsistency between $\widetilde{G_T}$ and $\widetilde{G}_S$ should be similar to the inconsistency between $\widehat{G_T}$ and $\widehat{G_S}$.

$\mu$ and $\lambda$ are important parameters which capture the importance of each term, and we will discuss these parameters in Section \ref{sec:alg}.
To optimize the model as in Eq.~\ref{equ:l2}, we use stochastic gradient descent~\cite{han2011data} method. The derivative on $u_T$ is given as:
\begin{equation}
\label{equ:dev}
\begin{split}
\frac{1}{2}\frac{\partial \mathcal{L}_{2}^{u_T}}{\partial E_T}
& = \mu u_T (u_Tu_T^T - \widetilde{A_T})  + (1 - \mu)u_T\left \|\frac{2\left\| u_Tu_T^T - \widetilde{A_S}\right\|}{ n_S(n_S-1)} - C_3 \right \| + u_T
\end{split}
\end{equation}

\subsection{Overall Algorithm}
\label{sec:alg}

\begin{algorithm}[t]
	\caption{\acret: Knowledge Transfer based Algorithm for Dependency Graph Learning}
    \label{alg:overall}
    \SetKwInOut{Input}{Input}
    \SetKwInOut{Output}{Output}

    \Input{ $G_S$, $\widehat{G_T}$}
    \Output{${G}_T$}
    Select a set of \textit{meta-path}s from $G_S$.\;
    Extract $|P|$ networks from $G_S$\;
    Calculate all the similarity matrix $S_{p_i}$\;
    
    {\color{gray} $\setminus *$ The Entity Estimation Process $ * \setminus$}\;
    \While{Convergence}
    {
        	Calculate $U_k$ and $\Lambda_k$\; 
    		$u_S = U_k\sqrt[2]{\Lambda_k}$.\;
        	Calculate $S_G$, $T$ and $C_1$\;
	        $W = (S_G^TS_G)^{-1}S_GC_1$.\;
    }     
    Construct $\widetilde{G_T}$ based on the method introduced in Section \ref{sec:mcc}\;
    
    {\color{gray} $\setminus *$ The Dependency Construction Process $ * \setminus$}\;
    \While{Convergence}
    {
    	Update $u_T$ using the gradient of function \ref{equ:dev};\  
    }
    Construct ${G}_T$ based on the method introduced in Section \ref{sec:msc};\ 
\end{algorithm}

The overall algorithm is then summarized as Algorithm \ref{alg:overall}.
\nop{This algorithm implements the methods we have introduced in the above subsections.}
In the algorithm, line 5 to line 11 implements the Entity Estimation Model, and line 13 to 16 implements the Dependency Construction Model.
\nop{In the result of this section, we will discuss some practical issues of the ACRET algorithm.}

\subsubsection{Setting Parameters}
There are two parameters, $\lambda$ and $\mu$, in our model. 
For $\lambda$, as in \cite{sun2012mining,han2011data}, it is always assigned manually based on the experiments and experience. \nop{Therefore, we only discuss the assignment of parameter $\mu$ in our model.} 
For $\mu$, when a large number of entities are transferred to the target domain, a large $\mu$ can improve the transferring result, because we need more information to be added from the source domain. 
On the other hand, when only a small number of entities are transferred to target domain, then a larger $\mu$ will bias the result. 
Therefore, the value of $\mu$ depends on how many entities are transferred from the source domain to the target domain.
In this sense, we can use the proportion of the transferred entities in $\widetilde{G}_T$ to calculate $\mu$. 
Given the entity size of $\widetilde{G}_T$ as $|\widetilde{G}_T|$, the entity size of $\widehat{G_T}$ as $|\widehat{G_T}|$, then $\mu$ can be calculated as:
\begin{equation}
\label{equ:para}
\mu = (|\widetilde{G_T}| - |\widehat{G_T}|)/|\widetilde{G_T}|,
\end{equation}




The experimental results in Section \ref{seq:para} demonstrate the effectiveness of the proposed parameter selection method.

\begin{figure*}[!htbp]
    \centering
    \begin{subfigure}[b]{0.33\textwidth}
        \includegraphics[width=\textwidth]{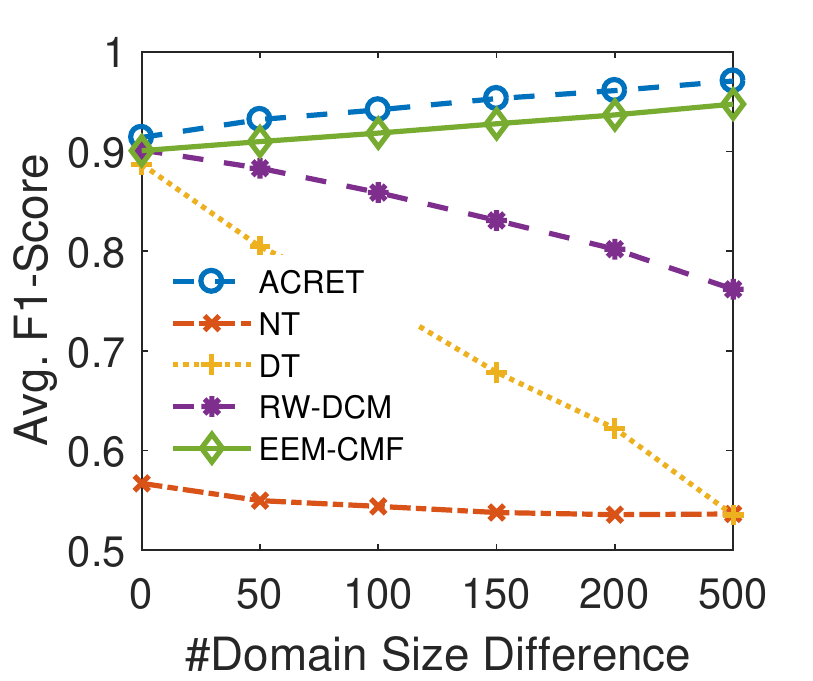}
        \caption{Varying graph size}
        \label{fig:modelsize}
    \end{subfigure}
    ~ 
    \begin{subfigure}[b]{0.33\textwidth}
        \includegraphics[width=\textwidth]{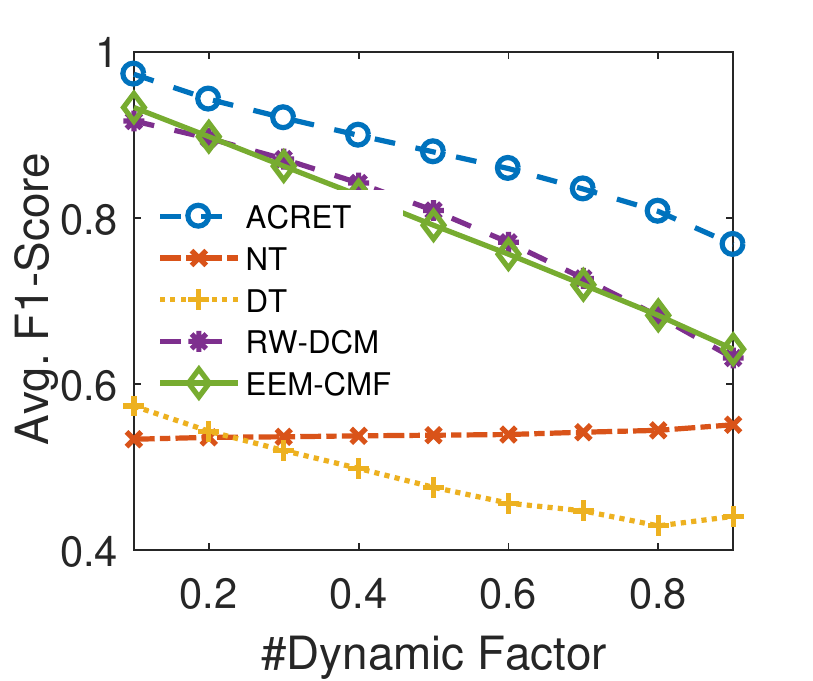}
        \caption{Varying dynamic factor}
        \label{fig:dynamicfactor}
    \end{subfigure}
    ~ 
    \begin{subfigure}[b]{0.33\textwidth}
        \includegraphics[width=\textwidth]{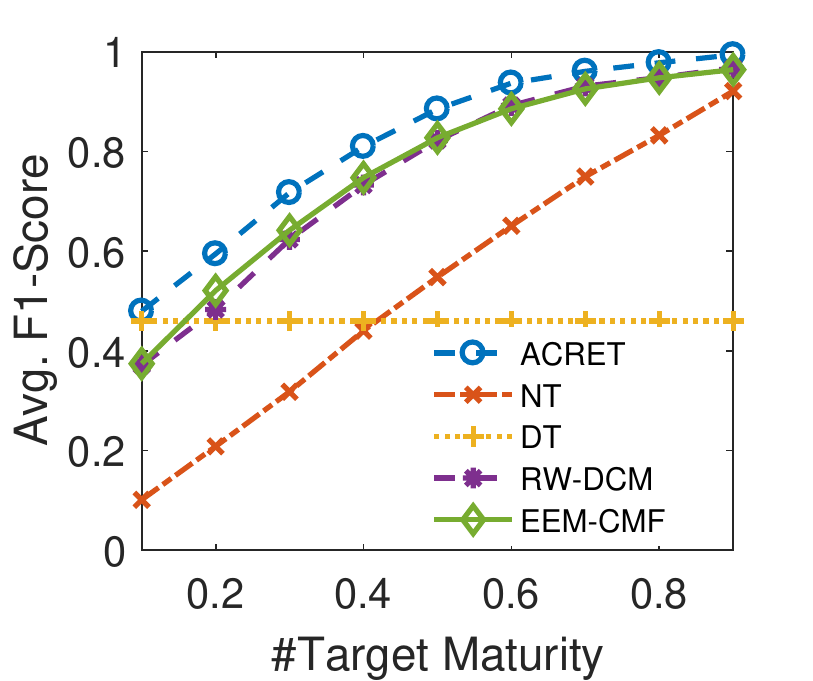}
        \caption{Vary graph maturity}
        \label{fig:modelmature}
    \end{subfigure}
    \caption{Performance on synthetic data.}
    \label{fig:synthetic}
\end{figure*}

\subsubsection{Complexity Analysis}
\label{sub:complexity}
As shown in Algorithm \ref{alg:overall}, the time for learning our model is dominated by computing the objective functions and their corresponding gradients against feature vectors. 

For the Entity Estimation Model, the time complexity of computing the $u_S$ in Eq.~\ref{equ:es} is bounded by $O(d_1n)$, where $n$ is the number of entities in $G_S$, and $d_1$ is the dimension of the vector space of $u_S$.
The time complexity for computing $W$ is also bounded by $O(d_1n)$.
So, suppose the number of training iterations for \textbf{EEM} is $t_1$, then the overall complexity of \textbf{EEM} model is $O(t_1d_1n)$. 
For the Dependency Construction Model, the time complexity of computing the gradients of $\mathcal{L}_{2}$ against $u_T$ is $O(t_2d_2n)$, where $t_2$ is the number of iterations, $d_2$ is the dimensionality of feature vector. 
As shown in our experiment (see Section \ref{sec:casecov}), $t_1$, $t_2$, $d_1$, and $d_2$ are all small numbers.
So that we can regard them as a constant, say $C$, so the overall complexity of our method is $O(Cm)$, which is linear with the size of the entity set. This makes our algorithm practicable for large scale datasets.

\section{Experiments}
\label{sec:exp}
In this section, we evaluate \acret\ using synthetic data and real system surveillance data collected in enterprise networks.

\subsection{Comparing Methods}
We compare \acret\ with the following methods:

\textbf{NT}:
This method directly uses the original small target dependency graph without knowledge transfer.
In other words, the estimated target dependency graph $G_T = \widehat{G_T}$.

\textbf{DT}:
This method directly combines the source dependency graph and the original target dependency graph.
In other words, the estimated target dependency graph $G_T= \widetilde{G_S} + \widehat{G_T}$. 

\textbf{RW-DCM}:
This is a modified version of the \acret\ method.
Instead of using the proposed \textbf{EEM} model to perform entity estimation, this method uses the random walk to evaluate the correlations between entities and perform entity estimation.
Random walk is a widely-used method for relevance search in a graph~\cite{kang2012fast}. \nop{For more details about random walk, please refer to \cite{kang2012fast}.}


\textbf{EEM-CMF}:
This is another modified version of the \acret\ method.
In this method, we replace \textbf{DCM} model with collective matrix factorization~\cite{singh2008relational} method.
Collective matrix factorization has been applied for link prediction in multiple domains \cite{singh2008relational}.
\nop{For the details about collective matrix factorization, please refer to \cite{singh2008relational}.}


\subsection{Evaluation Metrics}
Since in \acret\ algorithm, we use hypothesis-test for thresholding the selection of entities and dependencies, similar to \cite{luo2014correlating,han2011data}, we use the F1-score to evaluate the hypothesis-test accuracy of all the methods. 

\nop{Since all methods listed above are based on hypothesis test, similar to \cite{luo2014correlating,han2011data}, we use the F1-score to evaluate the hypothesis-test accuracy of all the methods. }
\nop{F1 score has been used in many research works for analyzing hypothesis-test accuracy.}
\nop{We use the F1-score to evaluate the performance of all the methods.} F1-score is the harmonic mean of precision and recall. In our experiment, the final F1-score\nop{between the estimated dependency graph $G_T$ and the ground-truth dependency graph $\overline{G_T}$} is calculated by averaging the entity F1-score and dependency/edge F1-score. 

To calculate the precision (recall) of both entity and link, we compare the estimated entity (edge) set with the ground-truth entity/link set. 
Then, precision and recall can be calculated as follows:
\[
Precision=\frac{N_{C}}{N_{E}},
{~~~~~~~}
Recall=\frac{N_{C}}{N_{T}},
\]
where $N_C$ is the number of correctly estimated entities (edges), $N_E$ is the number of total estimated entities (edges), and $N_T$ is the number of ground truth entities (edges). 


\subsection{Synthetic Experiments} 


We first evaluate the \acret\ on synthetic graph data-sets to have a more controlled setting for assessing algorithmic performance. 
We control three aspects of the synthetic data to stress test the performance of our \acret\ method: 
\begin{itemize}
\item \textbf{Graph size} is defined as the number of entities for a dependency graph. Here, we use $|G_S|$ to denote the source domain graph size and $|\widehat{G_T}|$ to denote the target one. 
\item \textbf{Dynamic factor}, denoted as $F$, has the same definition as in Section \ref{sec:link}. 
\item \textbf{Graph maturity score}, denoted as \textit{M}, is defined as the percentage of entities/edges of the ground-truth graph $\overline{G_T}$, that are used for constructing the original small graph $\widehat{G_T}$. Here, graph maturity score is used for simulating the period of learning time of $\widehat{G_T}$ to reach the maturity in the real system.
\end{itemize}

Then, given $|G_S|$, $|\widehat{G_T}|$, $F$, and \textit{M}, we generate the synthetic data as follows:
We first randomly generate an undirected graph as the source dependency graph $G_S$ based on the value of $|G_S|$ \cite{west2001introduction}; Then, we randomly assign three different labels to each entity. Due to space limitations, we will only show the results with three labels, but similar results have been achieved in graphs with more than three labels; 
We further construct the target dependency graph $\overline{G_T}$ by randomly adding/deleting $F = d\%$ of the edges and deleting $|G_S| - |\widehat{G_T}|$ entities from $G_S$.
Finally, we randomly select \textit{M} = $c\%$ of entities/edges from $\overline{G_T}$ to form $\widehat{G_T}$. 

\subsubsection{How Does \acret's Performance Scale with Graph Size?}
We first explore how the \acret's performance changes with graph size $|G_S|$ and $|\widehat{G_T}|$.
Here, we fix the maturity score to \textit{M}$=50\%$, the dynamic factor to $F=10\%$, and target domain dependency graph size to $|\widehat{G_T}|=0.9$.
Then, we increase the source graph size $|G_S|$ from $0.9K$ to $1.4K$.
From Fig. \ref{fig:modelsize}, we observe that with the increase of the size difference $|G_S| - |\widehat{G_T}|$, the performances of \textbf{DT} and \textbf{RW-DCM} are getting worse. This is due to the poor ability of \textbf{DT} and \textbf{RW-DCM} for extracting useful knowledge from the source domain.
In contrast, the performance of \acret\ and \textbf{EEM-CMF} increases with the size differences.
This demonstrates the great capability of \textbf{EEM} model for entity knowledge extraction. 
\nop{In sum, compared with all other methods, our \acret\ method achieves the best performance.}


\subsubsection{How Does \acret's Performance Scale with Domain Dynamic Factor?}
We now vary the dynamic factor $F$ to understand its impact on the \acret's performance. Here, the graph maturity score is set to \textit{M}=$50\%$, and two domain sizes are set to $|G_S|=1.2K$ and $|\widehat{G_T}|=0.6K$, respectively.
Fig. \ref{fig:dynamicfactor} shows that the performances of all the methods go down with the increase of the dynamic factor. This is expected, because transferring the dependency graph from a very different domain will not work well.
On the other hand, the performances of \acret, \textbf{RW-DCM}, and \textbf{EEM-CMF} only decrease slightly with the increase of the dynamic factor. Since \textbf{RW-DCM} and \textbf{EEM-CMF} are variants of the \acret\ method, this demonstrates that the two sub-models of the \acret\ method are both robust to large dynamic factors.
\subsubsection{How Does \acret's Performance Scale with Graph Maturity?}
Third, we explore how the graph maturity score \textit{M} impacts the performance of \acret.
Here, the dynamic factor is fixed to $F=0.2$.
The graph sizes are set to $|G_S| = 1.2K$ and $|\widehat{G_T}|=0.6$.
Fig. \ref{fig:modelmature} shows that with the increase of the \textit{M}, the performances of all the methods are getting better. The reason is straightforward: with the maturity score increases, the challenge of domain difference for all the methods is becoming smaller. 
In addition, our \acret\ and its variants \textbf{RW-DCM}, and \textbf{EEM-CMF} perform much better than \textbf{DT} and \textbf{NT}. This demonstrates the great ability of the sub-models of \acret\ for knowledge transfer.
Furthermore, \acret\ still achieves the best performance.

\subsection{Real-World Experiments}
\label{sub:real_data}

\begin{figure*}[t]
    \centering
    \begin{subfigure}[b]{0.33\textwidth}
        \includegraphics[width=\textwidth]{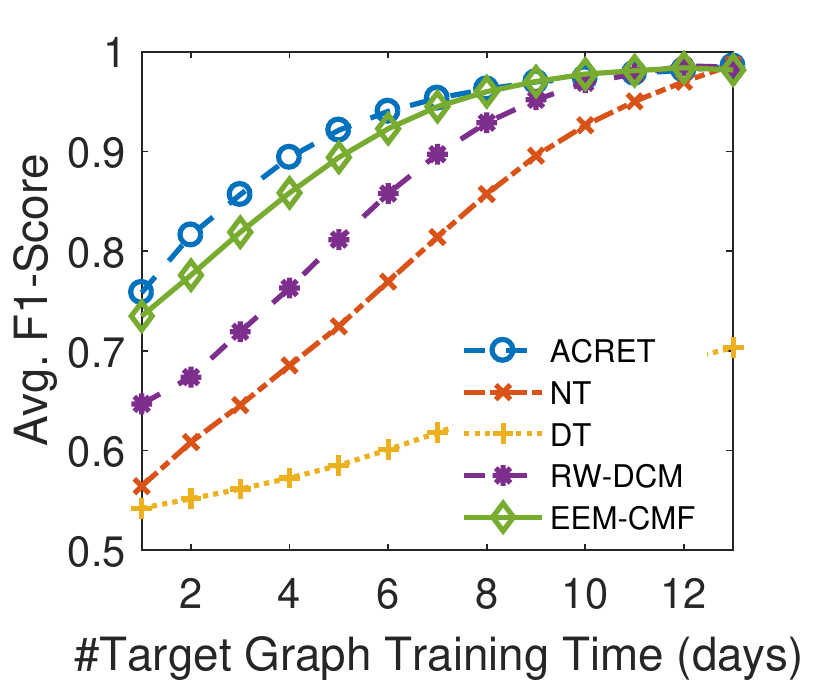}
        \caption{Windows dataset}
        \label{fig:win}
    \end{subfigure}
    ~ 
    \begin{subfigure}[b]{0.33\textwidth}
        \includegraphics[width=\textwidth]{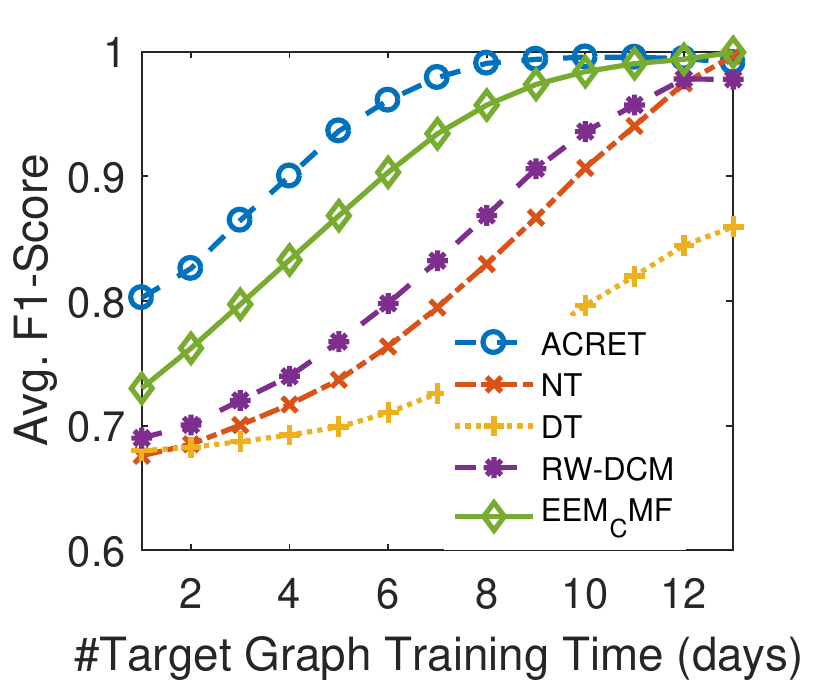}
        \caption{Linux dataset}
        \label{fig:linux}
    \end{subfigure}
    \caption{Performance on real-world data.}
    \label{fig:real}
\end{figure*}

Two real-world system monitoring datasets are used in this experiment.
The data is collected from an enterprise network system composed of $47$ Linux machines and $123$ Windows machines from two departments, in a time span of $14$ consecutive days.
In both datasets, we collect two types of system events: (1) communications between processes, and (2) system activity of processes sending or receiving Internet connections to/from other machines at destination ports.
Three different types of system entities are considered: (1) processes, (2)
Unix domain sockets, and (3) Internet sockets.
The sheer size of the Windows dataset is around $7.4$ Gigabytes, and the Linux dataset is around $73.5$ Gigabytes.
Both Windows and Linux datasets are split into a source domain and a target domain according to the department name.
The detailed statistics of the two datasets are shown in Table \ref{tab:realdata}.
\begin{table}[!htbp] 

 \caption{The statistics of real datasets.}

 \label{tab:realdata}
\centering
 \begin{tabular}{c|c|c} 
 \hline
 Data & Win & Linux \\
 \hline
 \# System Events & 120 Million & 10 Million \\
 \# Source Domain Machines & 62 & 24 \\
 \# Target Domain Machines & 61 & 23 \\
 \# Time Span & 14 days & 14 days\\
 \hline
 \end{tabular}

\end{table}

In this experiment, we construct one target domain dependency graph $\widehat{G_T}$ per day by increasing the learning time daily.
The final graph is the one learned for $14$ days.
\nop{The result is shown in Fig. \ref{fig:real}.}
From Fig. \ref{fig:real}, we observe that for both Windows and Linux datasets, with the increase of the training time, the performances of all the algorithms are getting better. 
On the other hand, compared with all the other methods, \acret\ achieves the best performance on both Windows and Linux datasets.
In addition, our proposed \acret\ algorithm can make the dependency graph deplorable in less than four days, instead of two weeks or longer by directly learning on the target domain.

\begin{figure}[!htbp]
    \centering
      \begin{subfigure}[b]{0.2\textwidth}
        \includegraphics[width=\textwidth]{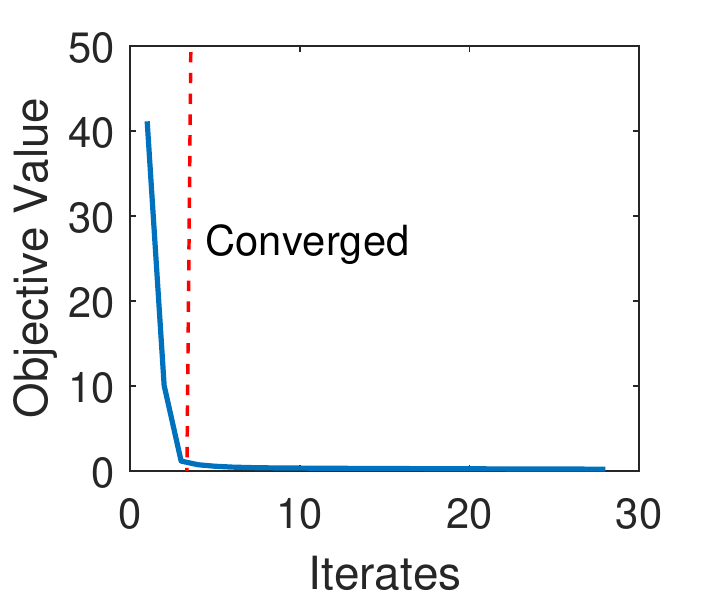}
        \caption{\textbf{EEM} on synthetic data}
        \label{fig:size_entity}
    \end{subfigure}
    \begin{subfigure}[b]{0.2\textwidth}
        \includegraphics[width=\textwidth]{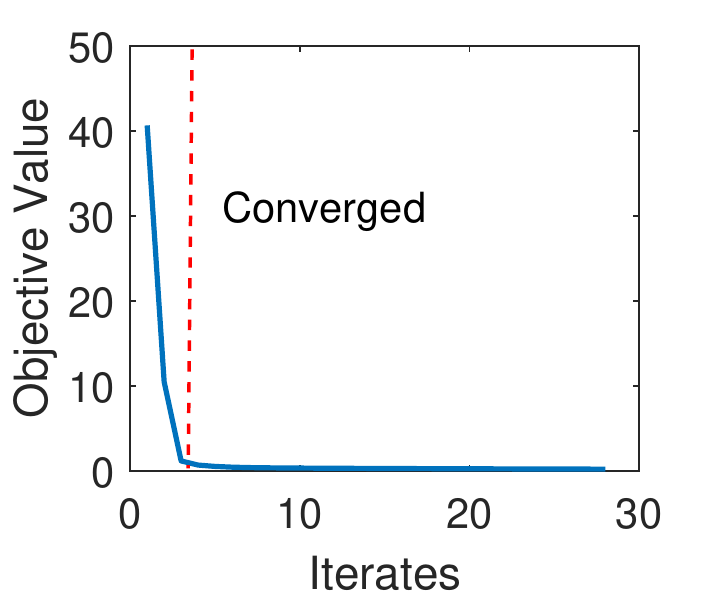}
        \caption{\textbf{DCM} on synthetic data}
        \label{fig:size_link}
    \end{subfigure}
    \begin{subfigure}[b]{0.2\textwidth}
        \includegraphics[width=\textwidth]{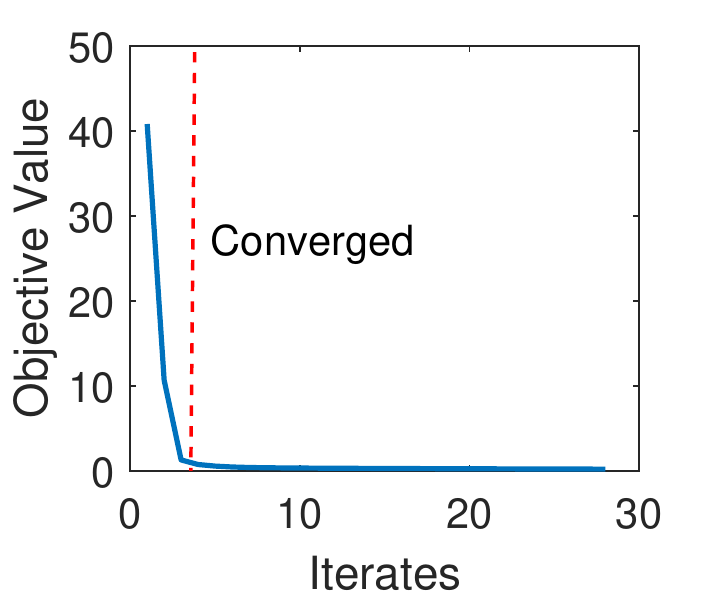}
        \caption{\textbf{EEM} on Windows data}
        \label{fig:size_entity}
    \end{subfigure}
    \begin{subfigure}[b]{0.2\textwidth}
        \includegraphics[width=\textwidth]{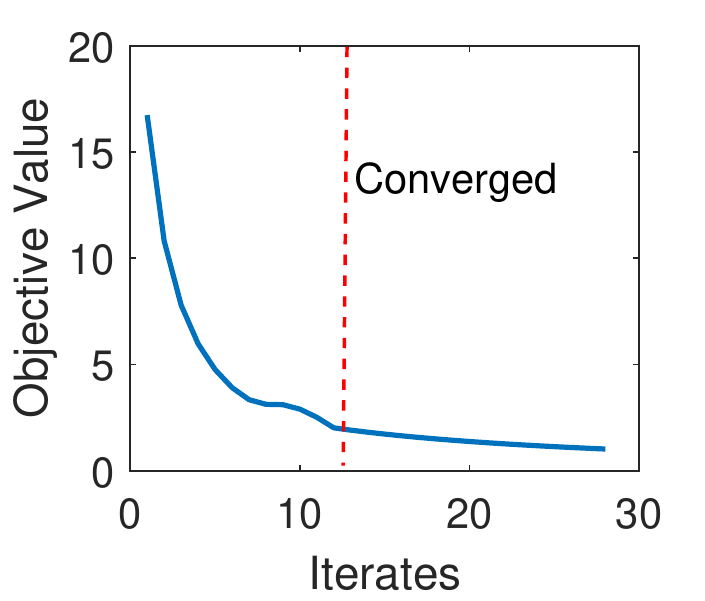}
        \caption{\textbf{DCM} on Windows data}
        \label{fig:size_link}
    \end{subfigure}    
    \begin{subfigure}[b]{0.2\textwidth}
        \includegraphics[width=\textwidth]{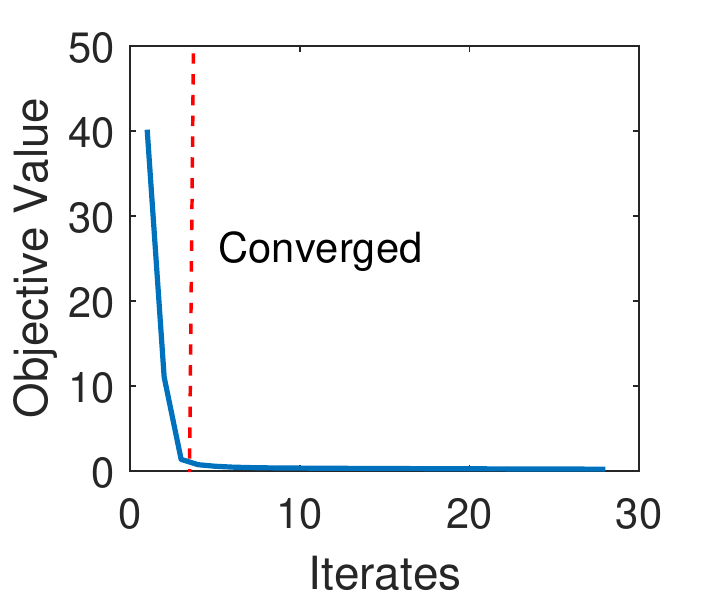}
        \caption{\textbf{EEM} on Linux data}
        \label{fig:size_entity}
    \end{subfigure}
    \begin{subfigure}[b]{0.2\textwidth}
        \includegraphics[width=\textwidth]{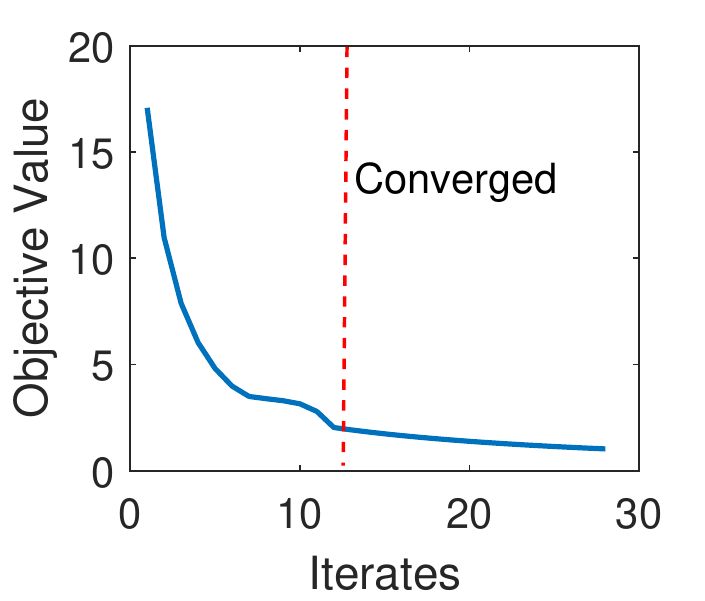}
        \caption{\textbf{DCM} on Linux data}
        \label{fig:size_link}
    \end{subfigure} 
    \caption{Performance on convergence.}
    \label{fig:convrate}
    
\end{figure}

\subsection{Convergence Analysis}
\label{sec:casecov}
As described in Section \ref{sub:complexity}, the performance bottleneck of \acret\ model is the learning process of the two sub-models: \textbf{EEM} (Entity Estimation Model) and \textbf{DCM} (Dependency Construction Model). 
In this section, we report the convergence speed of our approach.

We use both synthetic and real-world data to validate the model convergence speed.
For the synthetic data, we choose the one with dynamic factor to be $F = 0.2$, the dependency graph size to be $|G_S|=1.2K$ and $|\widehat{G_T}|=0.6K$, and the graph maturity to be $50\%$.
For the two real-world datasets, we fix the target dependency graph learning time as $4$ days. \nop{The convergence rate of different models is shown in Fig. \ref{fig:convrate}.}

From Fig. \ref{fig:convrate}, we can see that in all three datasets, \acret\ converges very fast (\textit{i.e.}, with less than 10 iterations). This makes our model applicable for the real-world large-scale systems.

\subsection{Parameter Study}
\label{seq:para}

\begin{figure*}[!htbp]
    \centering
    \begin{subfigure}[b]{0.30\textwidth}
        \includegraphics[width=\textwidth]{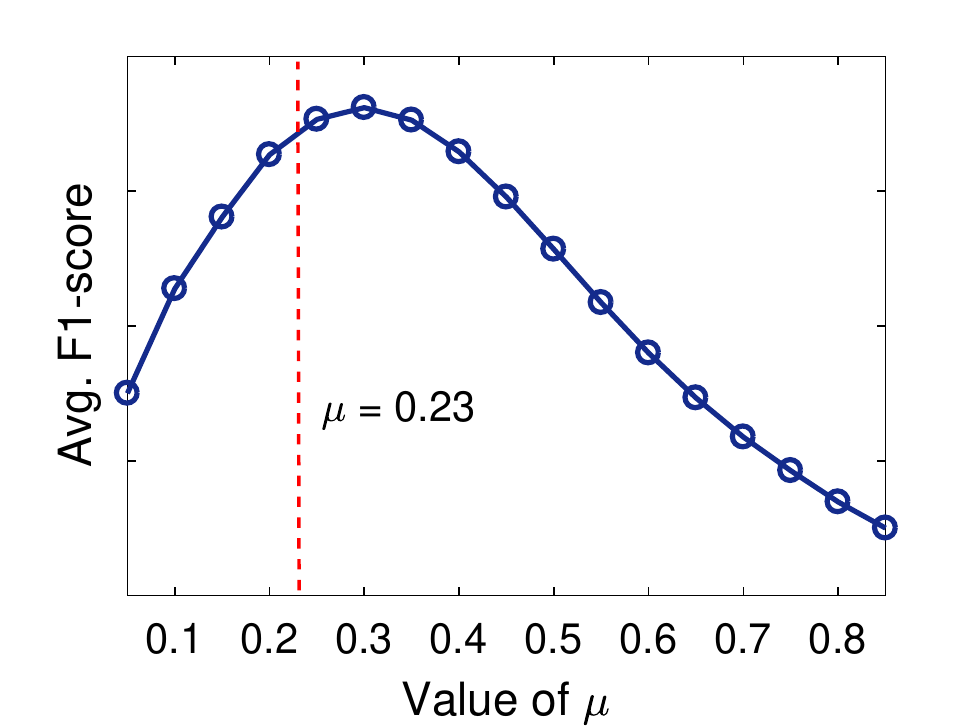}
        \caption{Synthetic dataset}
        \label{fig:size_entity}
    \end{subfigure}
    ~ 
    \begin{subfigure}[b]{0.30\textwidth}
        \includegraphics[width=\textwidth]{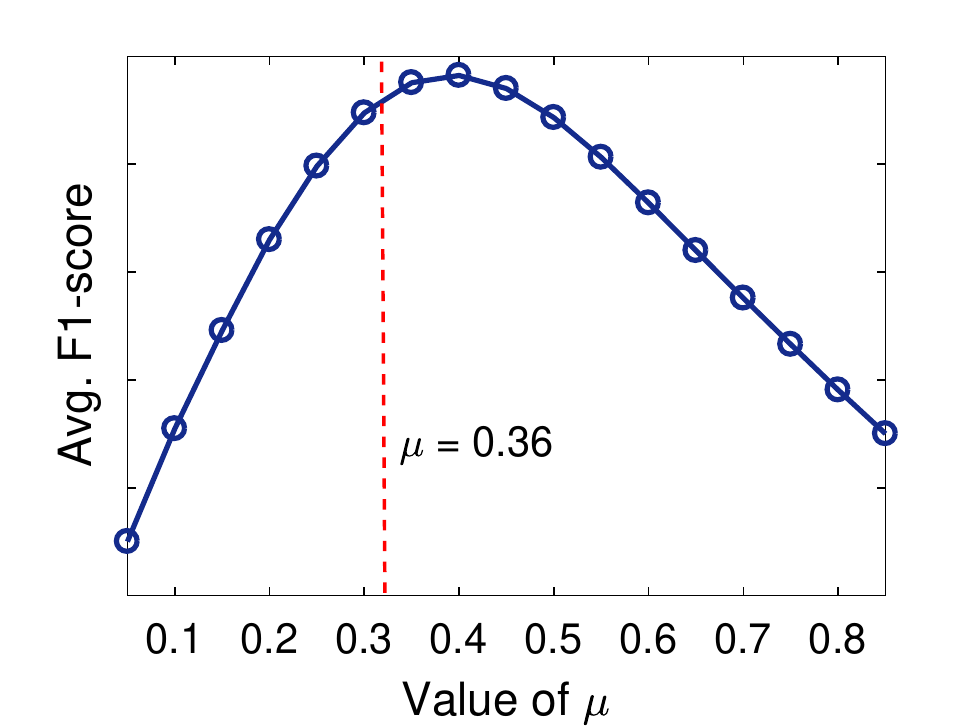}
        \caption{Windows dataset}
        \label{fig:size_link}
    \end{subfigure}
    \begin{subfigure}[b]{0.30\textwidth}
        \includegraphics[width=\textwidth]{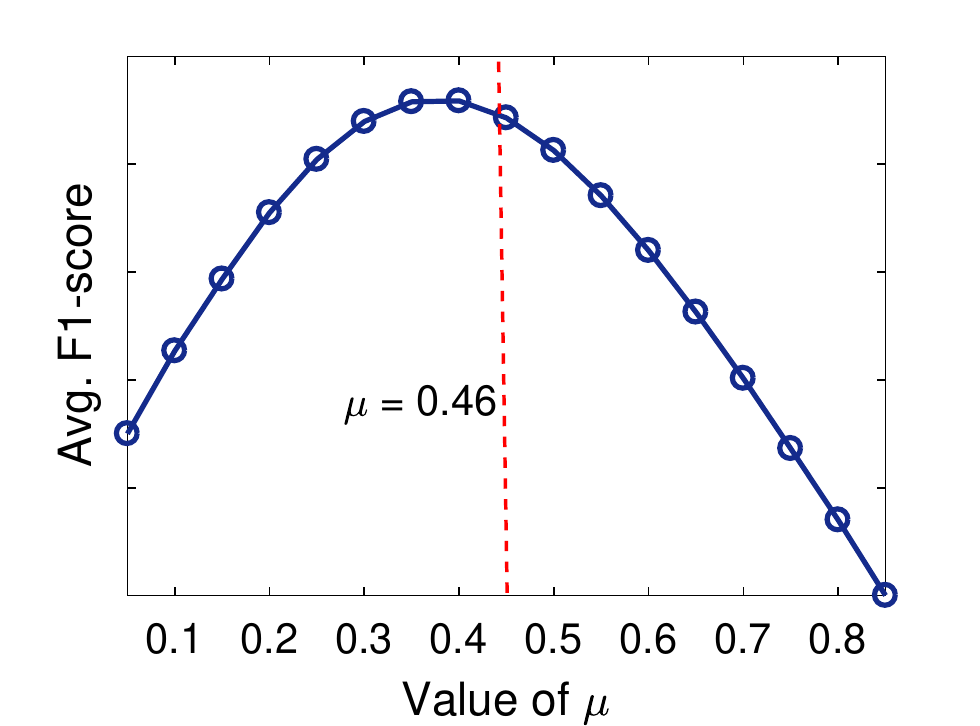}
        \caption{Linux dataset}
        \label{fig:size_entity}
    \end{subfigure} 
    \caption{Sensitivity analysis on parameter $\mu$.}
    \label{fig:parameter}
\end{figure*}

\nop{The result is shown in Fig. \ref{fig:parameter}.}
In this section, we study the impact of parameter $\mu$ in Eq.~\ref{equ:l2}.
We use the same datasets as in Section \ref{sec:casecov}. As shown in Fig. \ref{fig:parameter}, when the value of $\mu$ is too small or too large, the results are not good, because $\mu$ controls the leverage between the source domain information and target domain information.
The extreme value of $\mu$ (too large or too small) will bias the result.
On the other hand, the $\mu$ value calculated by Eq. \ref{equ:para} is $0.23$ for the synthetic dataset, $0.36$ for the Windows dataset, and $0.46$ for the Linux dataset. And Fig. \ref{fig:parameter} shows the best results just appear around these three values.
This demonstrates that our proposed method for setting the $\mu$ value is very effective, which successfully addresses the parameter pre-assignment issue.

\subsection{Case Study on Intrusion Detection}
As aforementioned, dependency graph is essential to many forensic analysis applications like root cause diagnosis and risk analysis. In this section, we evaluate the \acret's performance in a real commercial enterprise security system (see Fig.~\ref{fig:motivating}) for intrusion detection. 

In this case, the dependency graph, which represents the normal profile of the enterprise system, is the core analysis model for the offshore intrusion detection engine. It is built from the normal system process event streams (see Section \ref{sub:real_data} for data description) during the training period.\nop{The security system has just been deployed in one Japanese electric company for three days, but it has been deployed in a US IT company for 1 month. In order to achieve good intrusion detection results in the electric company with the only three days' training, \acret\ is applied for accelerating the dependency graph learning process by leveraging the knowledge learned from the IT company.} The same security system has been deployed in two companies: one Japanese electric company and one US IT company. We obtain one dependency graph from the IT company after 30 days' training, and two dependency graphs from the electric company after 3 and 30 days' training, respectively. \acret\ is applied for leveraging the well-trained dependency graph from the IT company to complete the 3 days' immature graph from the electric company. \nop{in the Electronic company with the only three days' training, \acret\ is applied for accelerating the dependency graph learning process by leveraging the knowledge learned from the IT company.}

In the one-day testing period, we try $10$ different types of attacks \cite{jones2000computer}, including Snowden attack, ATP attack, botnet attack, Sniffer Attack and \textit{etc.}, which resulted in $30$ ground-truth alerts. All other alerts reported during the testing period are considered as false positives.

\nop{We deploy \acret\ to a real world commercial anomaly detection system in an IT company X in Princeton, NJ. 
The anomaly detection system in this company utilizes the dependency graph as the core analysis engine.
Learning the dependency graph for the detection system needs more than three weeks.}
\nop{We omit and anonymize some details for confidentiality reasons.}

\nop{Table \ref{tab:application} shows the intrusion detection results using the dependency graph generated by \acret, comparing to other four baseline methods.} 

Table \ref{tab:application} shows the intrusion detection results in the electric company using the dependency graphs generated by different transfer learning methods and the 30 days' training from the electric company. From the results, we can clearly see that \acret\ outperforms all the other transfer learning methods by at least $18\%$ in precision and $13\%$ in recall. On the other hand, the performance of the dependency graph ($3$ days' model) accelerated by \acret\ is very close to the ground truth model ($30$ days' model). This means, by using \acret, we can achieve similar performance in one-tenth training time, which is of great significant to some mission critical environments.

\begin{table}[!htpb] 
 \caption{Intrusion detection performance.}
 \label{tab:application}
\centering
 \begin{tabular}{c|c|c} 
 \hline
 Method & Precision & Recall \\
 \hline
 \textbf{NT} & 0.01 & 0.10 \\
 \hline
 \textbf{DT} & 0.15  & 0.30 \\
 \hline
 \textbf{RW-DCM} &0.38   & 0.57\\
 \hline
 \textbf{EEM-CMF} &0.42  & 0.60\\
  \hline
  \acret\ & \textbf{0.60}  & 0.73 \\
 \hline
 \hline
\textbf{Real 30 days' model} & 0.58  & \textbf{0.76} \\
 \hline
 \end{tabular}
\end{table}

\nop{
\In this case study experiment, we choose the anomaly detection system of an electronic factory Y in XXX as the source domain.
We choose the the branch office of X in Princeton as the target domain.
The target domain original graph is a $3$ day immature dependency graph.
We deployed \acret\ and the comparing methods into the anomaly detection system of X.
The experimental result of anomaly detection test is reported in Table \ref{tab:application}. The real target model in Table. \ref{tab:application} denotes the model trained in Y for two weeks.
From the result, we can clearly see that \acret\ outperforms all the other methods.
On the other hand, the performance of the dependency graph ($3$ days model) accelerated by \acret\ is very close to the ground truth model ($30$ days model).
That means, by using \acret, we can achieve the same performance in less than half training time, which is of great significant to industry companies. }

\section{Related Work}
\label{sec:related}
\nop{In this section, we briefly introduce some related research efforts.}
\subsection{Transfer Learning}
Transfer learning has been widely studied in recent years \cite{cao2010transfer,pan2010survey}. Most of the traditional transfer learning methods focus on numerical data~\cite{dai2007boosting,sun2015transfer,chattopadhyay2013joint}. 
When it comes to graph (network) structured data, there is less existing work.
\nop{One line of transfer learning research that related to our work is transfer learning on the graph (network) structured data.}
In \cite{fang2015trgraph}, the authors presented \textit{TrGraph}, a novel transfer learning framework for network node classification. \textit{TrGraph} leverages information from the auxiliary source domain to help the classification process of the target domain.\nop{A similar approach proposed by them in \cite{fang2013transfer} is \cite{fang2013transfer}. In~\cite{fang2013transfer}, the authors proposed a transfer learning method} In one of their earlier work, a similar approach was proposed \cite{fang2013transfer} to discover common latent structure features as useful knowledge to facilitate collective classification in the target network. In~\cite{he2009graph}, the authors proposed a framework to propagates the label information from the source domain to the target domain via the example-feature-example tripartite graph.
Transfer learning has also been applied to the deep neural network structure. 
In \citep{chen2015net2net}, the authors introduced \textit{Net2Net}, a technique for rapidly transferring the information stored in one neural net into another. \textit{Net2Net} utilizes function preserving transformations to transfer knowledge from neural networks. Different from existing methods, we aim to expedite the dependency graph learning process through knowledge transfer. 
\nop{However, none of the methods have done the knowledge transfer for reconstruction the dependency graphs.}

\subsection{Link Prediction and Relevance Search}
Graph link prediction is a well-studied research topic~\cite{liben2007link, hofman2017prediction}.
\nop{There are some link prediction works that related to our paper.} 
In \cite{ye2013predicting}, Ye \textit{et al.} presented a transfer learning algorithm to address the edge sign prediction problem in signed social networks. Because edge instances are not associated with a pre-defined feature vector, this work was proposed to learn the common latent topological features shared by the target and source networks, and then adopt an AdaBoost-like transfer learning algorithm with instance weighting to train a classifier. 
\nop{The other work related to our work is collective matrix factorization} Collective matrix factorization \cite{singh2008relational} is another popular technique that can be applied to detect mission links by combining the source domain and target domain graphs. However, all the existing link prediction methods can not deal with dynamics between the source domain and target domain as introduced in our problem.
 
\nop{\textbf{Graph Relevance Search}}
Finding relevant nodes or similarity search in graphs is also related to our work. Many different similarity metrics have been proposed such as Jaccard coefficient, cosine similarity, and Pearson correlation coefficient \cite{bondy1976graph}, and Random Walks \cite{kang2012fast, sun2005relevance}. 
However, none of these similarity measures consider the multiple relations exist in the data.
Recent advances in heterogeneous information networks \cite{sun2012mining} have offered several similarity measures for heterogeneous relations, such as meta-path and relation path \cite{luo2014hete,luo2014hetpathmine}.
However, these methods can not deal with the multiple domain knowledge.

\section{Conclusion}
\nop{Dependency graphs capture intrinsic relationships between different pairs of system entities of physical systems. A well-trained dependency graph can be used for many data analysis application such as anomaly detection, system behavior analysis, recommendation, \textit{etc}.
However, learning dependency graph is a time-consuming process.
To speed up the learning process, 
In this paper, we proposed \acret, the first transfer learning model for accelerating dependency graph learning. 
\acret\ can effectively extract useful knowledge (\textit{e.g.}, entity and dependency relations) from the source domain, and transfer it to the target dependency graph. 
\acret\ also  the dynamic factor between different domains.
The experimental results on simulation datasets and two real datasets demonstrate the effectiveness and efficiency of \acret.}
\nop{We also deployed ACRET into real world commercial systems.}

In this paper, we investigate the problem of transfer learning on dependency graph. Different from traditional methods that mainly focus on numerical data, we propose \acret, a two-step approach for accelerating dependency graph learning from heterogeneous categorical event streams. 
By leveraging entity embedding and constrained optimization techniques, \acret\ can effectively extract useful knowledge (\textit{e.g.}, entity and dependency relations) from the source domain, and transfer it to the target dependency graph. 
\acret\ can also adaptively learn the differences between two domains, and construct the target dependency graph accordingly.
We evaluate the proposed algorithm using extensive experiments. The experiment results convince us of the effectiveness and efficiency of our approach.
We also apply \acret\ to a real enterprise security system for intrusion detection.
Our method is able to achieve superior detection performance at least 20 days lead lag time in advance with more than 70\% accuracy.

\bibliographystyle{ACM-Reference-Format}
\bibliography{modeltransfer} 

\end{document}